\documentclass[conference]{IEEEtran}
\IEEEoverridecommandlockouts

\usepackage{amsmath,amssymb,amsfonts}
\usepackage{algorithmic}
\usepackage{graphicx}
\usepackage{textcomp}
\usepackage{xspace}
\usepackage{xcolor}
\usepackage{booktabs}
\usepackage[numbers]{natbib}
\usepackage{url}
\usepackage{caption}
\usepackage{tikz}
\usepackage{amsmath}

\usepackage{subcaption}
\usepackage{pgfplots}
\pgfplotsset{compat=newest}
\pgfplotsset{major grid style={black!20}}

\usepackage{cleveref}

\def\BibTeX{{\rm B\kern-.05em{\sc i\kern-.025em b}\kern-.08em
    T\kern-.1667em\lower.7ex\hbox{E}\kern-.125emX}}

\newcommand{\xai}{XAI\xspace}
\newcommand{\ie}{i.e.\xspace}
\newcommand{\shap}{\textsc{Shap}\xspace}
\newcommand{\gradcam}{\textsc{Grad-CAM}\xspace}
\newcommand{\rex}{\textsc{ReX}\xspace}
\newcommand{\lime}{\textsc{Lime}\xspace}
\newcommand{\deeplift}{\textsc{DeepLIFT}\xspace}
\newcommand{\kernelshap}{\textsc{KernelShap}\xspace}
\newcommand{\gradientshap}{\textsc{GradientShap}\xspace}
    
\begin{document}

\title{Quantifying Explainable AI-introduced signal noise on ECG data with Spectral Entropy}

\author{\IEEEauthorblockN{David A. Kelly}
\IEEEauthorblockA{\textit{Dept. of Informatics} \\
\textit{King's College London}\\
London, UK \\
david.a.kelly@kcl.ac.uk}
\and
\IEEEauthorblockN{Nathan Blake}
\IEEEauthorblockA{\textit{Dept. of Informatics} \\
\textit{King's College London}\\
London, UK \\
nathan.blake@kcl.ac.uk}
}

\maketitle

\begin{abstract}
Explainability techniques are used to assess the output of various deep learning models. This is especially true in
healthcare, where models need to be trusted and decisions justified. Explainability (\xai) tools use heuristics which often add signal noise to the explanation ``core''. It is not always obvious what is signal from the
model and what is noise from the \xai. We propose the use of spectral entropy as a measure of noise in \xai output. We
demonstrate its usefulness in the context of classifying arrhythmias in an ECG dataset with different post hoc
explainability techniques. 
\end{abstract}


\section{Introduction}
\begin{figure}[t]
    \centering
    \includegraphics[scale=0.5]{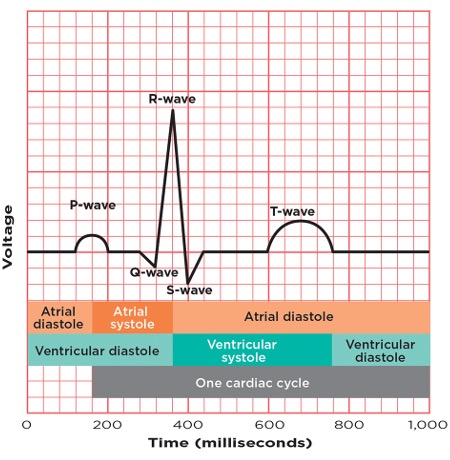}
    \caption{A single ECG complex. Schematic taken from \cite{jarvis2021electrocardiogram}}
    \label{fig:ecg_complex}
\end{figure}

An ECG assesses the electrophysiology of the heart in a quick and non-invasive manner. The x-axis comprises time,
usually expressed as milliseconds, and voltage in millivolts on the y-axis. As seen in~\Cref{fig:ecg_complex}, it
comprises several prominent features which map to specific phases of the cardiac cycle. For instance, the P-wave
indicates atrial systole, the QRS complex ventricular systole and the T-wave repolarization. Hence, it is possible to
diagnose multiple pathologies from an ECG.

The development of wearable devices has made possible the prospect of remote ECG monitoring. This could help facilitate
effective triage and the appropriate allocation of health resources. However, this requires the AI model to be small
enough to fit onto such a device. If an anomaly is detected, this would need to be verified by an expert to reduce the
risk of a false positive and the inappropriate allocation of health resources. Alternatively, false negatives indicate
an undetected arrhythmia. This could also lead to the inappropriate allocation of finite health resources. \xai
facilitates quick and effective verification by drawing an expert's attention to the features which contributed the most
to the model's classification, as long as the \xai itself does not introduce too much ``noise'' by highlighting
irrelevant features not actually used by the model.

Multiple \xai tools have been used to explain the outputs of AI arrhythmia detectors, with \shap~\citep{shap} and
\gradcam~\citep{gradcam} being particularly common choices~\citep{salih2024review}. However, nearly half of all
published \xai applications do not evaluate the explanations themselves~\citep{salih2024review}. This omission is even
more common in the medical domain~\cite{jung2023essential}, where expert knowledge is at a premium and frequently not
available for large scale studies. Recent guidelines, called FUTURE-AI, for the development and deployment of AI tools
in a medical context recommend utilizing \xai~\cite{lekadir2025future}. This includes, among other recommendations,
quantitative evaluations and measuring sensitivity to noise of the \xai tool itself (as distinct from the AI model). The
difficulty of scaling the assessment of \xai tools together with the development of clinical guidelines, such as
FUTURE-AI, has given rise to a growing interest in establishing proxy-measures for quantitatively assessing the outputs
of \xai.

In this paper, we treat the output of the \xai tool itself as a signal. This signal is a combination of \emph{model
signal} (the actual features used by the model for classification) and \xai-introduced noise. All post hoc
explainability tools are heuristic based: \shap, \rex~\citep{CKKS24,kelly2025causal} and \lime~\citep{lime} for example, rely on some
form of sampling, and \lime also uses segmentation information for images. These heuristics inevitably introduce some
noise into the underlying model signal, which we will call \emph{self-noise}. A good \xai tool should correctly identify
model signal, but should ideally introduce the smallest amount of self-noise, or -- failing that -- predictable
self-noise. We propose using spectral entropy as a means to quantify how much self-noise \xai tools add to the signal.
To the best of our knowledge, no one has proposed a means to systematically examine self-noise before. We evaluate
spectral entropy to quantify self-noise on ECG data.

\xai output is not time-series; we are using spectral entropy in an unusual way. This paper demonstrates that this
slight abuse is justified by practical use. Spectral entropy can augment other \xai assessment methods (see
\Cref{sec:discussion}), but does not answer all questions. Entropy can measure the self-noise of the \xai tool but
cannot be used directly to assess the ``accuracy'' of an explanation. Indeed, it is an open question whether a model's
criteria for classification should align with a human expert's criteria~\cite{bhusalface}, although in the short-term,
clinical applications will need to align with prior medical knowledge~\cite{jin2023guidelines}. However, spectral
entropy can be used to measure the ``precision'' of an explanation. An explanation with low accuracy, \ie in the wrong
place, may still have high precision due to its concentration or lack of diffusion. This high precision can be captured
by spectral entropy. 

\paragraph*{Qualities of a good explanation}~\citet{jin2023guidelines} specify that explanations should be
understandable to end users, clinically relevant, agree with prior medical (informative plausibility) knowledge and
truthfulness. The former requires that \xai outputs are concise and presented coherently. Indeed, one of the principal
complaints from medical users of \xai is the amount of superfluous signal ~\citep{alattal2025integrating,
bienefeld2023solving}. Spectral entropy quantifies this intuition. The latter requirement is common to all similar
recommendations and necessitates that the \xai tool is assessed for the degree to which it captures what the AI model is
actually doing. This would include assessing whether the \xai tool adds self-noise due to its heuristics.

\section{Spectral Entropy}
Shannon entropy is a measure of the average surprisal in a system~\citep{CoverThomas2006}. A system in which every outcome is equally surprising is a maximum entropy system. In \xai output, maximum entropy would be achieved by uniform-at-random allocation of importance to the different input features (\Cref{fig:pvc_ks}). This would be akin to white noise in signal processing. Any \xai tool which outputs white noise would be maximally `bad`, on the assumption that the explanation itself if not uniformly distributed over the feature space (see \Cref{sec:discussion}). We use the \emph{scipy} implementation of the \emph{periodogram} function to compute the power spectral density of the explanation (\Cref{eq:1}). 

\begin{equation}\label{eq:1}
    P(\omega_i) = \frac{1}{N}\mid X(\omega_i)\mid^2
\end{equation}

This is then normalized (\Cref{eq:2}) and passed directly to the standard formula for Shannon entropy, to give us the power spectral entropy $PSE = - \sum_{i=1}^{n} p_i \log p_i$. Such a calculation is inexpensive to perform in real time. All results are in information theoretic \emph{bits}, i.e. the $\log$ is base $2$.

\begin{equation}\label{eq:2}
    p_i = \frac{P(\omega_i)}{\sum_i P(\omega_i)}
\end{equation}

\xai explanations obviously do not have a sample rate. The \emph{scipy} implementation of the periodogram function
assumes a default sample rate of $1$. We find in our evaluation that changing the sample rate has no effect on the
calculation of spectral entropy. 

\section{Experiments}

To demonstrate the usefulness of this measure we utilized a model trained to detect cardiac arrhythmias from ECG data.
\citet{sun2021arrhythmia} specifically developed a model small enough to be contained within a wearable device. Their 17
layer, 1D-convolutional neural network was trained on the MIT-BIH dataset~\citep{moody2001impact} to classify 17
arrhythmias to an accuracy of $93.5$\%. As the model from~\citet{sun2021arrhythmia} was not available, we trained a
model ourselves from their instructions.

We used all $1000$ ECG from this dataset to classify and then explain the model predictions using \gradcam, \lime,
\deeplift~\citep{deeplift} and \rex~\cite{CKKS24}, specifically its version for spectral data~\cite{blake2025specrex}.
We also used two variations of \shap: \gradientshap and \kernelshap. We utilized a \lime variant designed for time
series data~\citep{sivill2022limesegment}, rather than standard \lime. Apart from this variant and
\rex\footnote{https://github.com/ReX-XAI/ReX}, all \xai tools were obtained from the
\emph{captum.ai}\footnote{\url{https://captum.ai/}} library. \gradcam directly accesses the final convolutional layer of
a model, therefore its output is the same size as this final layer, which in this model results in an activation map of
6 features. This is then upsampled via linear interpolation into the original $3600$ features.

All tools were used with their default settings. Such is the parameter space of each tool that it would be unfeasible to
explore these spaces adequately. The techniques run the gamut from white box, \ie \gradcam with its direct access to
convolutional layers, to purely black box, \ie \rex, which relies only on the model classification. All the \xai
techniques provide post hoc explanations of a particular instance of an ECG. The model that we use is not inherently
explainable, so we were unable to estimate the entropy of any self-explanatory method.

The output of each tool is not directly comparable. \rex and \lime, for example, do not compute values below $0$,
whereas tools which estimate Shapley values produce both positive and negative values. Negative values indicate features
which either ``distract'' from the actual classification, or point towards another classification. For the sake of fair
comparison, we ignore sub-zero values when calculating the spectral entropy of a given explanation as we are only
interested in features which positively contribute to the classification. We additionally took ECGs from the PVC
(Premature Ventricular Contraction) class and annotated a ground truth explanation based on domain expertise. Such
manual annotation is notoriously expensive: we annotated 20 ECGs from this class. From this, we calculate a ground truth
entropy for this class. 

\begin{figure*}[t]
    \centering 
    \begin{subfigure}{0.3\textwidth}
        \scalebox{0.6}{
        \begin{tikzpicture}
            \begin{axis}[grid=major, xlabel=Hz]
                \addplot table [x=x, y=y, col sep=comma, mark=none] {images/gs_psd.csv};
            \end{axis}
        \end{tikzpicture}
        }
        \caption{\gradientshap entropy $9.0$}\label{fig:psd_gs}
     \end{subfigure}
     \hfill
     \begin{subfigure}{0.3\textwidth}
        \scalebox{0.6}{
        \begin{tikzpicture}
            \begin{axis}[grid=major, xlabel=Hz]
                \addplot table [x=x, y=y, col sep=comma, mark=none] {images/ks_psd.csv};
            \end{axis}
        \end{tikzpicture}
        }
        \caption{\kernelshap entropy $10.19$}\label{fig:psd_ks}
     \end{subfigure}
      \hfill
     \begin{subfigure}{0.3\textwidth}
        \scalebox{0.6}{
        \begin{tikzpicture}
            \begin{axis}[grid=major, xlabel=Hz]
                \addplot table [x=x, y=y, col sep=comma, mark=none] {images/lime_psd.csv};
            \end{axis}
        \end{tikzpicture}
        }
        \caption{\lime entropy $3.73$}\label{fig:psd_lime}
     \end{subfigure}
     \caption{Example power spectral density (y-axis) plots for some of the \xai tools. The spread of noise is surprisingly wide, especially for \kernelshap, which
     seems fundamentally unsuited to this type of data and model.}%
     \label{fig:psd}
\end{figure*}
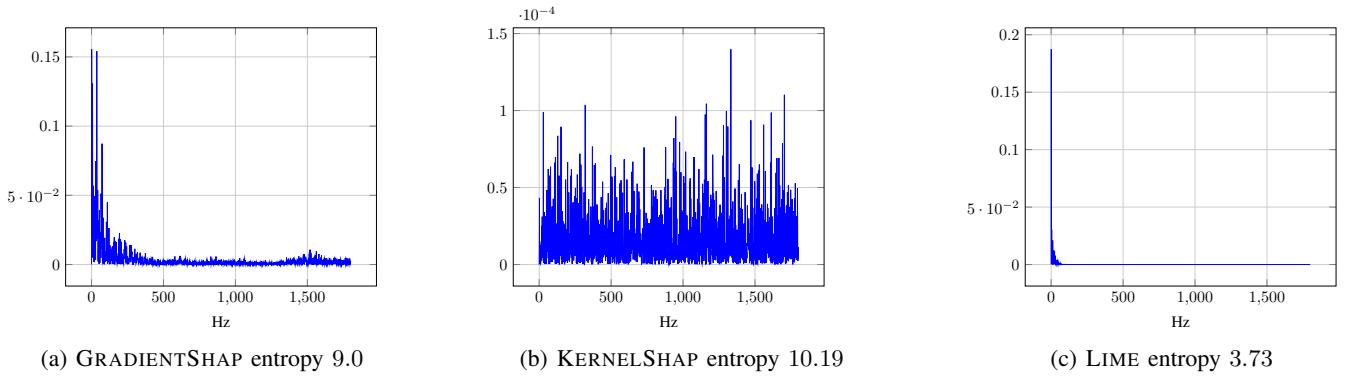

\section{Results}

\begin{table}[t]
    \centering
     \begin{tabular}{l|rr}
    \toprule\toprule
    Method        & Mean & Std \\
    \midrule
    \gradientshap  & 9.16   & 0.33    \\
    \kernelshap    & 10.21   & 0.04    \\
    \deeplift      & 9.04   & 0.39    \\
    \gradcam       & 1.07       & 0.58    \\
    \lime          & 10.20    & 0.02    \\
    \rex           & 1.93 & 1.20 \\
    \bottomrule
\end{tabular}
\caption{Average Spectral Entropy (+/- 1 SD) by \xai tool}%
\label{tab:entropy_results}
\end{table}

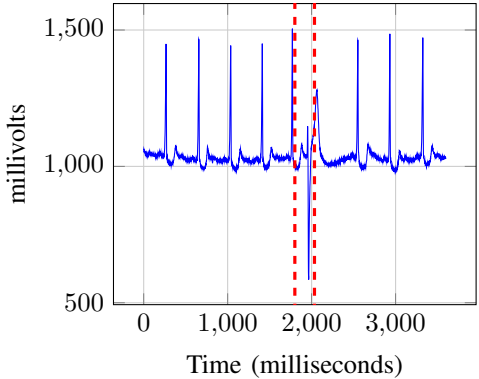
\begin{figure}[t]
    \centering
    \begin{tikzpicture}
      \pgfplotsset{scale=0.7}
          \begin{axis}[grid=major, xlabel=Time (milliseconds), ylabel=millivolts]
                 \addplot table [x=x, y=y, col sep=comma, mark=none] {images/raw.csv};
                 \draw [dashed, very thick, color=red] (1800, 0) -- (1800, 1600);
                 \draw [dashed, very thick, color=red] (2034, 0) -- (2034, 1600);
             \end{axis}
       \end{tikzpicture}
     \caption{PVC ECG with approximate location marked between red lines.}%
     \label{fig:results_visual}
\end{figure}

\Cref{tab:entropy_results} shows the average spectral entropy for each tool across all $17$ classes. The most obvious feature is that, for most tools, the entropy does not vary much over all $17$ classes. This is a good indicator that, for this model and data, most tools produce a near constant amount of self-noise. The exception to this is \rex, which has the highest standard deviation. Unlike the other tools, \rex is capable of identifying multiple explanations~\citep{CKKmultiple25} in images. It may be the case that the feature attribution map output by \rex contains multiple explanations for certain classes, resulting in the wide spread of entropy values. This needs further investigation.

\begin{figure*}[t]
    \centering 
    \begin{subfigure}{0.3\textwidth}
        \scalebox{0.65}{
        \begin{tikzpicture}
            \begin{axis}[grid=major, xlabel=Time (milliseconds)]
                \addplot table [x=x, y=y, col sep=comma, mark=none] {images/rex.csv};
                \draw [dashed, very thick, color=red] (1800, 0) -- (1800, 100);
                \draw [dashed, very thick, color=red] (2034, 0) -- (2034, 100);
            \end{axis}
        \end{tikzpicture}
        }
        \caption{\rex}\label{fig:pvc_rex}
     \end{subfigure}
     \hfill
     \begin{subfigure}{0.3\textwidth}
        \scalebox{0.65}{
        \begin{tikzpicture}
            \begin{axis}[grid=major, xlabel=Time (milliseconds)]
                \addplot table [x=x, y=y, col sep=comma, mark=none] {images/gs.csv};
                \draw [dashed, very thick, color=red] (1800, 0) -- (1800, 1);
                \draw [dashed, very thick, color=red] (2034, 0) -- (2034, 1);
            \end{axis}
        \end{tikzpicture}
        }
        \caption{\gradientshap}\label{fig:pvc_gs}
     \end{subfigure}
     \hfill
     \begin{subfigure}{0.3\textwidth}
        \scalebox{0.65}{
        \begin{tikzpicture}
            \begin{axis}[grid=major, xlabel=Time (milliseconds)]
                \addplot table [x=x, y=y, col sep=comma, mark=none] {images/ks.csv};
                \draw [dashed, very thick, color=red] (1800, 0) -- (1800, 0.1);
                \draw [dashed, very thick, color=red] (2034, 0) -- (2034, 0.1);
            \end{axis}
        \end{tikzpicture}
        }
        \caption{\kernelshap}\label{fig:pvc_ks}
     \end{subfigure}
     \begin{subfigure}{0.3\textwidth}
        \scalebox{0.65}{
        \begin{tikzpicture}
            \begin{axis}[grid=major, xlabel=Time (milliseconds)]
                \addplot table [x=x, y=y, col sep=comma, mark=none] {images/dl.csv};
                \draw [dashed, very thick, color=red] (1800, 0) -- (1800, 1);
                \draw [dashed, very thick, color=red] (2034, 0) -- (2034, 1);
            \end{axis}
        \end{tikzpicture}
        }
        \caption{\deeplift}\label{fig:pvc_dl}
     \end{subfigure}
     \hfill
     \begin{subfigure}{0.3\textwidth}
        \scalebox{0.65}{
        \begin{tikzpicture}
            \begin{axis}[grid=major, xlabel=Time (milliseconds)]
                \addplot table [x=x, y=y, col sep=comma, mark=none] {images/gc.csv};
                \draw [dashed, very thick, color=red] (1800, 0) -- (1800, 2.3);
                \draw [dashed, very thick, color=red] (2034, 0) -- (2034, 2.3);
            \end{axis}
        \end{tikzpicture}
        }
        \caption{\gradcam}\label{fig:pvc_gc}
     \end{subfigure}
     \hfill
     \begin{subfigure}{0.3\textwidth}
        \scalebox{0.65}{
        \begin{tikzpicture}
            \begin{axis}[grid=major, xlabel=Time (milliseconds)]
                \addplot table [x=x, y=y, col sep=comma, mark=none] {images/lime.csv};
                \draw [dashed, very thick, color=red] (1800, 0) -- (1800, 0.1);
                \draw [dashed, very thick, color=red] (2034, 0) -- (2034, 0.1);
            \end{axis}
        \end{tikzpicture}
        }
        \caption{\lime}\label{fig:pvc_lime}
     \end{subfigure}
     \caption{Explanations from different \xai tools for the PVC in~\Cref{fig:results_visual}. The approximate ground truth is indicated between the red dotted lines.  Units on the different $y$-axes are incomparable, of importance is relative magnitude; the $x$-axis is time. \rex (\Cref{fig:pvc_rex}), \gradientshap (\Cref{fig:pvc_gs}) and \deeplift (\Cref{fig:pvc_dl}), on this example, give most weight to the ground truth location. \gradcam is in the general area, while \lime highlights an incorrect region. \kernelshap output is close to white noise.}%
     \label{fig:explanations}
\end{figure*}
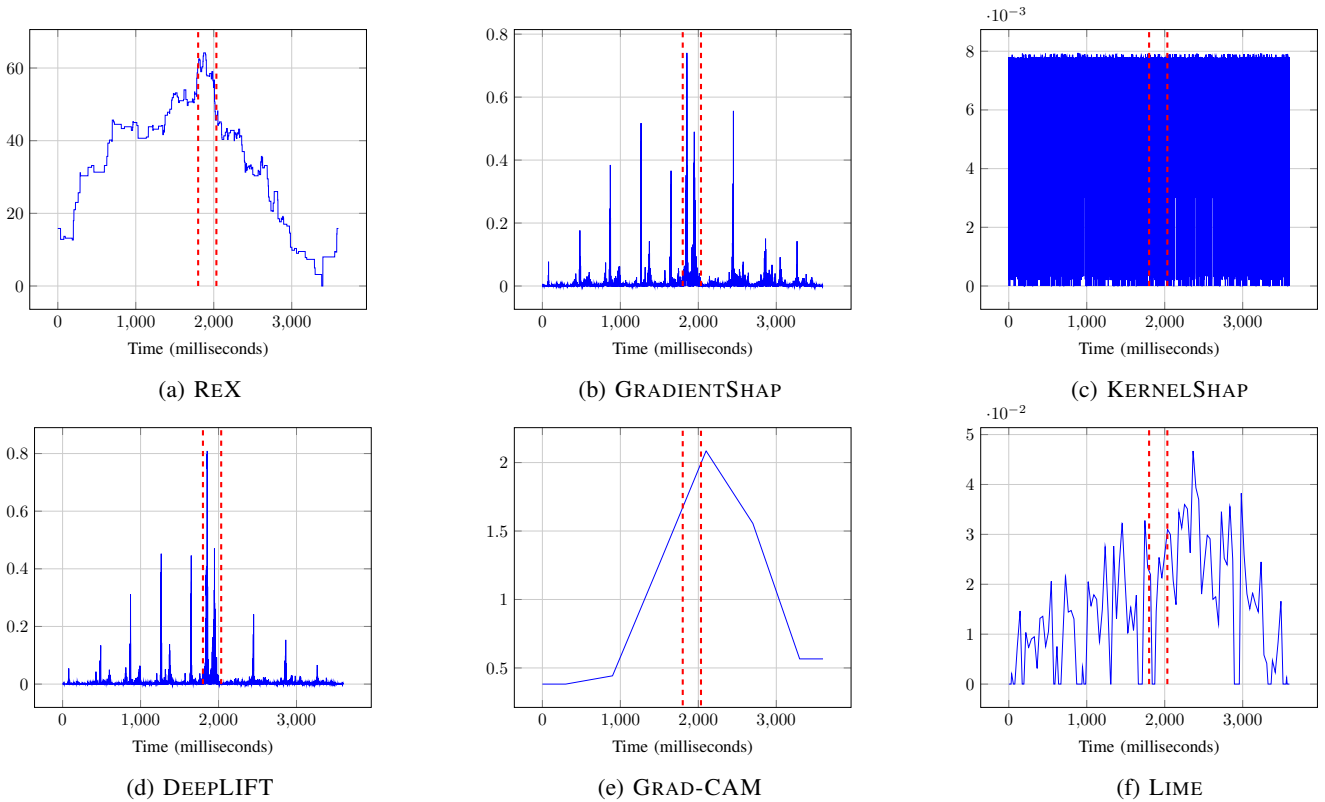

We focus in on one class to show the utility of the method. \Cref{fig:results_visual} shows an ECG with a PVC (premature ventricular contraction) highlighted, representing the only region required for a human domain expert to classify it. This particular example has only one explanation: the region between the dotted red lines. \Cref{fig:explanations} shows the output of the different \xai tools on this ECG instance. The appropriate region has again been highlighted in red.

The most obvious difference is between tools which estimate Shapley values, and those which do
not.~\Cref{fig:pvc_gs,fig:pvc_dl,fig:pvc_ks} all estimate Shapley values. \kernelshap has a signal which is close to
pure noise (\Cref{fig:psd_ks}). One would have to assume that \kernelshap is a poor choice of tool for ECG
explainability. Both \gradientshap and \deeplift highlight the correct region in that their highest peaks are within the
red zone. However, they have other, clearly separate, peaks which are spurious. Indeed, \gradientshap has a low level of
noise which is fairly constant (\Cref{fig:psd_gs}). \rex output is different, showing a gradual rise towards the model
signal and then a fall away. \gradcam shows a similar pattern, though with almost no self-noise. It does not, however,
highlight the correct part of the signal. This may be due to the necessity to upsample \gradcam's output. In lieu of
more detailed information, this upsampling is uniform, assigning each of the $6$ neurons to equal width bins in the
output space. It may be that equal binning is not the most appropriate method, but it is unclear what one might do in
its place. \lime output is curious, showing a relatively small amount of noise (\Cref{fig:psd_lime}) while still failing
to highlight the correct part of the signal.

\section{Discussion}\label{sec:discussion}
The tools fall broadly into two classes: high noise and low noise. This does not necessarily align with accuracy of
explanation. For instance, \lime has relatively low entropy across all classes, but on PVC at least does not accurately
select any of the appropriate peaks. Conversely, although \gradientshap has high entropy on the same signal, the largest
peaks are clearly aligned with the relevant parts of the ECG for PVC. While this level of accuracy does not generalize
over all classes, the noise does: spectral entropy is fairly consistent over all classes for all tools, with the
exception of \rex. This suggests that it is accurately capturing the baseline heuristic noise of each tool.

An important assumption when discussing model signal and self-noise is that the explanation (or rather, the data that
the model is using to make its classification) is local, \ie confined to just a few points in the signal. A non-local
explanation, where the explanation is distributed across the entire input, might indeed look close to white noise. A low
spectral entropy will be unfortunate in the event that an explanation is non-local, and therefore the signal really is
evenly distributed over the input space. This is not the case with the ECG data that we use, however.

Various metrics have been suggested to quantify the quality of \xai output. \citet{petsiuk2018rise} introduce deletion
and insertion curves, which track model confidence as features are ablated or introduced. \citet{fidelity} propose the
concepts of \emph{sensitivity} and \emph{fidelity}, which measure the changes in an explanation due to perturbations on
the original input. All of these methods test the resilience or portability of an explanation in some way, by removing
or introducing features, or perturbing feature values. All of these methods require multiple calls to a model and are
computationally expensive.

None of these metrics attempt to quantity how much noise the \xai tool itself adds into the explanation. \citet{CKKS24}
use the number of pixels in an explanation of an image as a quality measure. They find that the Shapley value based
tools require more pixels in general than either \rex or \lime. This is likely analogous to self-noise, whereby
unnecessary pixels are added to an explanation which do not actually contribute to the class. They do not, however,
attempt to quantity how much spurious information is added by the tools they investigate.

\citet{miller2019explanation} argues that an explanation in the medical context is inherently social. Any quantitative
performance measure is a proxy measure of its usefulness. This meta-analysis problem is well documented in the
literature~\citep{hedstrom2023meta}. Spectral entropy is an elegant technique which measures the degree of spurious
information the \xai method adds, aligning with the qualitative feeling of an output containing ``too much'' to be
useful.

\section{Conclusion}
We have proposed the use of spectral entropy as an \xai evaluation metric for medical, tabular-style, data. It
quantifies the intuition that \xai outputs with spurious information obscures domain expert interpretation. Spectral
entropy can be used alongside other quantitative measures when assessing the \xai tool which seeks to explain the
outputs of an ECG classifier. Spectral entropy is computationally inexpensive and can be used in real time. This would
not be instead of domain expert assessment, but may be helpful when scaling such assessments to a quantity beyond their
resource limitations. 

\section*{Acknowledgments}
David Kelly and Nathan Blake are funded by CHAI: EPSRC - Causality in Healthcare AI Hub (grant number EP/Y028856/1).

\bibliographystyle{IEEEtranN}
\bibliography{all}

\end{document}